%% file: report_template.tex
\documentclass[11pt]{article}

\usepackage[letterpaper,margin=1in]{geometry}
\usepackage[T1]{fontenc}
\usepackage[utf8]{inputenc}
\usepackage{lmodern}
\usepackage{microtype}
\usepackage{booktabs}
\usepackage{longtable}
\usepackage{array}
\usepackage{graphicx}
\usepackage{tabularx}
\usepackage{enumitem}
\usepackage[table]{xcolor}
\usepackage[numbers,sort&compress]{natbib}
\usepackage{soul}
\usepackage[hidelinks]{hyperref}
\usepackage[most]{tcolorbox}
\usepackage{titlesec}
\usepackage{caption}
\usepackage{fancyhdr}
\usepackage{float}
\usepackage{wrapfig}

\newcommand{\evalrx}{\textsc{EvalRx}}

\newfloat{algorithm}{tbp}{loa}
\floatname{algorithm}{Algorithm}
\floatstyle{plaintop}
\restylefloat{algorithm}
\newtcolorbox{principlebox}{
  colback=evblue!3,
  colframe=evblue!65!black,
  boxrule=0.8pt,
  arc=2pt,
  left=7pt,
  right=7pt,
  top=6pt,
  bottom=6pt
}
\definecolor{improvementgreen}{HTML}{16803A}
\newcommand{\improve}[1]{%
  \textcolor{improvementgreen}{$\Delta\,{+}#1\%$}%
}
\definecolor{evblue}{RGB}{29,78,216}
\definecolor{evnavy}{RGB}{15,32,61}
\definecolor{evslate}{RGB}{71,85,105}
\definecolor{evteal}{RGB}{11,110,110}
\definecolor{evgreen}{RGB}{31,122,61}
\definecolor{evred}{RGB}{176,58,46}
\definecolor{evborder}{RGB}{194,204,199}
\newcommand{\tableheader}{\rowcolor{evblue!12}}
\newcommand{\tablebanding}{\rowcolors{2}{evblue!3}{white}}

\titleformat{\section}
  {\Large\bfseries\color{evblue}}{\thesection}{0.6em}{}
  [\vspace{0.20em}\color{evblue!35}\titlerule]
\titleformat{\subsection}
  {\large\bfseries\color{evblue!85!black}}{\thesubsection}{0.6em}{}
  
\titlespacing*{\subsection}{0pt}{1.45em}{0.45em}
\renewcommand{\headrulewidth}{0.45pt}
\renewcommand{\headrule}{\hbox to\headwidth{\color{evblue!55}\leaders\hrule height \headrulewidth\hfill}}

\makeatletter
\renewcommand{\maketitle}{%
  \vspace*{-4.1em}
  \noindent
  {\small\sffamily\bfseries\color{evblue}\textls[70]{TECHNICAL REPORT}}%
  \hfill{\small\color{evslate}\@date}\par
  \vspace{0.42em}
    {\raggedright\noindent\fontsize{27}{29}\selectfont\sffamily\bfseries\color{evblue}
  \textls[88]{EvalRx}\par}
  \vspace{0.05em}
  {\raggedright\fontsize{18}{20}\selectfont\bfseries\color{black}
    \noindent Agentic Diagnosis and Verified Repair\par
    \noindent of Foundation Models\par}
  \vspace{0.22em}
  {\raggedright\noindent\normalsize\color{evslate}
    From model evaluation to statistically verified improvements across text, vision, and audio\par}
  \vspace{0.55em}
  \begin{tcolorbox}[
    colback=evblue!2,
    colframe=evblue!22,
    boxrule=0.5pt,
    arc=1mm,
    left=10pt,right=10pt,top=7pt,bottom=7pt,
    before skip=0pt,after skip=0pt
  ]
    \centering
    {\footnotesize\sffamily\bfseries\color{evblue}\textls[75]{EVALVITALS TEAM}\par}
    \vspace{5pt}
    {\small\color{black!75}\@author\par}
  \end{tcolorbox}
  \vspace{0.65em}%
}
\makeatother

\title{\evalrx{}: Closing the Loop from Model Evaluation to Auditable Repair Recipes [-2pt]\large An Agentic System for Auditable Diagnosis and Repair Across Modalities}
\newcommand{\preprintauthors}{%
  {\footnotesize\bfseries\color{evslate}Leading contributors}\enspace
  Jiaqi Liu\textsuperscript{1,2,*}\quad
  Ruinan Jin\textsuperscript{1,2,*}\quad
  Xiaoxiao Li\textsuperscript{1,2,*,\(\dagger\)}\par
  {\footnotesize\color{evslate}
    \href{mailto:jiaqiliu@ece.ubc.ca}{jiaqiliu@ece.ubc.ca}\quad
    \href{mailto:ruinanjin@alumni.ubc.ca}{ruinanjin@alumni.ubc.ca}\quad
    \href{mailto:xiaoxiao.li@ece.ubc.ca}{xiaoxiao.li@ece.ubc.ca}\par}
  \vspace{4pt}
  {\footnotesize\bfseries\color{evslate}Core contributors}\enspace
  Yitong Tang\textsuperscript{1}\quad
  Beidi Zhao\textsuperscript{1,2}\quad
  Gexin Huang\textsuperscript{1,2}\quad
  Qiong Zhang\textsuperscript{3}\par
  \vspace{4pt}
  {\footnotesize\color{evslate}
    \textsuperscript{1}The University of British Columbia\quad
    \textsuperscript{2}Vector Institute\par
    \textsuperscript{3}Mohamed bin Zayed University of Artificial Intelligence (MBZUAI)\par
  }
  \vspace{3pt}
  {\footnotesize\color{evslate}
    \textsuperscript{*}Equal contribution. Authors are listed in alphabetical order.\par
    \textsuperscript{\(\dagger\)}Corresponding author: Xiaoxiao Li.\par
  }%
}
\author{\preprintauthors}
\date{September 2026}

\begin{document}
\maketitle
\thispagestyle{empty}

\begin{tcolorbox}[
    enhanced,
    colback=evblue!3,
    colframe=evblue,
    boxrule=0pt,
    leftrule=3pt,
    arc=0pt,
    left=8pt,
    right=8pt,
    top=5pt,
    bottom=5pt,
    before skip=0pt,
    after skip=5pt
]
{\footnotesize\sffamily\bfseries\color{evblue}\textls[60]{ABSTRACT}}\par\vspace{2pt}
{\footnotesize
Model evaluation reveals \emph{where} models fail, but rarely \emph{why} or whether a proposed fix genuinely works. We introduce \evalrx{}, an open-source agentic framework for statistically grounded model diagnosis and repair. \evalrx{} discovers systematic failure modes, explores candidate mechanisms, formulates falsifiable hypotheses, and verifies them by executing frozen, pre-specified tests before using supported diagnoses to guide tiered repair. Candidate repairs are validated against the unchanged model and promoted only when they improve the target behavior without unacceptable regression. Across language, vision-language, and audio-language settings spanning Gemini, Qwen, Gemma, and Nemotron, \evalrx{} improves performance in all 12 evaluated model family–modality combinations, with gains ranging from \(+1.4\) to \(+10.9\) percentage points (\(+5.8\) points on average). These results show that model evaluation can serve not only to measure failures, but to support an auditable process for diagnosing and repairing them.
}
\end{tcolorbox}

\par\vspace{-5pt}
{\centering
  \includegraphics[
    width=0.9\textwidth,
    trim={0cm 2.5cm 0cm 2.5cm},
    clip
  ]{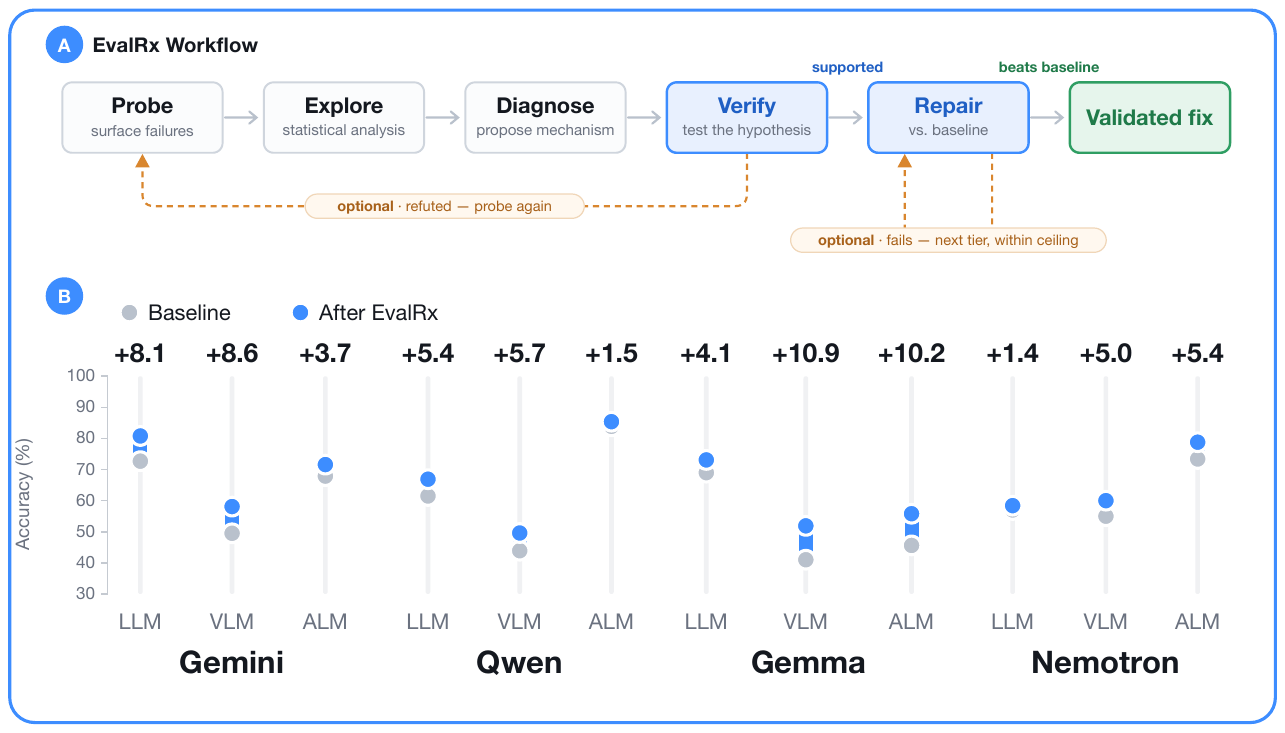}\par
  \vspace{-20pt}
  {\scriptsize\color{evslate}\textbf{
    \evalrx{} automatically boosts the model performance
    across diverse LLMs, VLMs and ALMs.
  }\par}
}
\par\vspace{4pt}


{\centering
\footnotesize\color{evslate}
\href{https://github.com/evalvitals/evalrx}
     {Code and artifacts: github.com/evalvitals/evalrx}
\quad$\cdot$\quad
\href{https://evalvitals.github.io/evalvitals-pages/}
     {Project overview}
\quad$\cdot$\quad
\href{https://evalvitals.github.io/evalvitals-pages/report/}
     {Demo}
\par}




\clearpage
\input{sections/introduction}
\input{sections/relatedwork}

\clearpage
\input{sections/method}
\input{sections/setup}
\input{sections/experiment}
\input{sections/conclusion}

\clearpage
\markboth{References}{References}
\bibliographystyle{plain}
\bibliography{sections/reference}
\clearpage
\appendix
\markboth{Appendix}{Appendix}
\input{sections/appendix}

\end{document}

%% file: sections/introduction.tex
\section{Introduction}
AI agents are becoming the execution layer for software engineering, data analysis, and scientific work. Systems built on foundation models (FMs) already resolve software issues autonomously \citep{jimenez2024swebench}, execute machine-learning engineering pipelines end-to-end~\citep{chan2025mlebench}, and participate in scientific research~\citep{majumder2024discoverybench,lu2024aiscientist,gottweis2026coscientist}. However, an agent is only as reliable as its behavior in the particular conditions it encounters. A model can possess the capacity required for a task while systematic, localized errors prevent that capacity from being expressed reliably, under a prompt perturbation, for a visual relation, or on a long-tail entity class. Unlike intrinsic capacity limits, these operational failure modes can be observed, diagnosed, and sometimes repaired.

This creates a deployment-critical gap. Diagnosing failures should work like a routine \ul{\emph{model health check}}: localize where errors concentrate, formulate an actionable explanation, and verify that an intervention improves the affected behavior without causing unacceptable regressions. Aggregate benchmarks reveal that a model fails, but not where, why, or whether an apparent fix generalizes. Expert review of individual errors is costly and hard to scale. LLM- or agent-based inspection is scalable, but can produce fluent conclusions unsupported by valid statistical evidence \citep{baumann2025llmhacking,asher2026phack}. \emph{The problem is structural: a workflow that searches many slices and explanations will find appealing post-hoc patterns that do not survive a pre-registered test with multiplicity control, or repairs that do not survive confirmation on unseen cases.}

\begin{principlebox}
\centering
\textbf{\evalrx{} is an open-source agentic system for closed-loop model diagnosis and repair, turning failure analysis into statistically verified model improvements.}
\end{principlebox}

\begin{wrapfigure}{r}{0.43\textwidth}
\vspace{-6pt}
\centering
\includegraphics[width=0.42\textwidth]{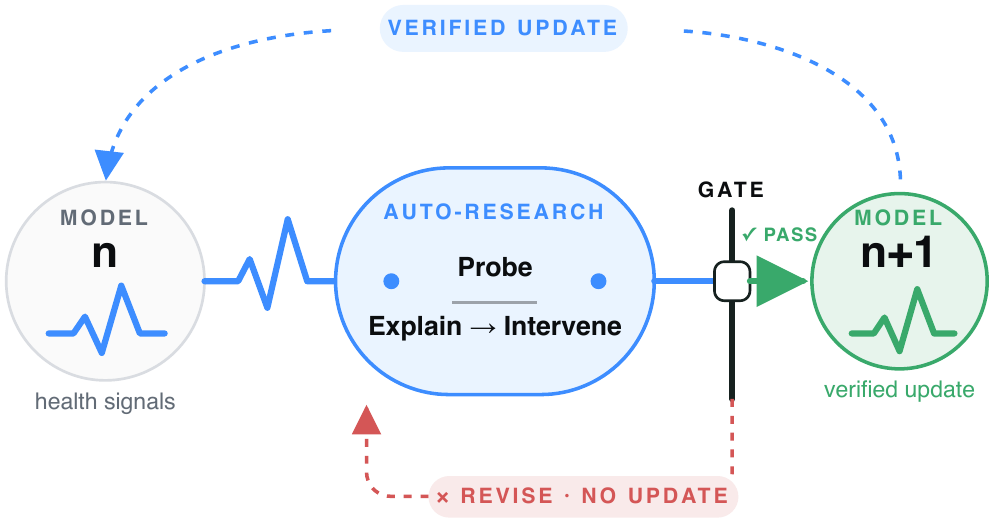}
\caption{\textbf{Verification-gated model evolution.} A candidate repair is promoted to the next model version only after its improvement is confirmed on held-out data. Otherwise, the current model remains unchanged and the diagnosis-repair cycle continues.}
\label{fig:recursive-teaser}
\vspace{-6pt}
\end{wrapfigure}

It closes the missing evidence-to-repair loop (Figure~\ref{fig:recursive-teaser}): starting from a fixed labeled evaluation set, it discovers candidate failure modes from case-level evidence, freezes a falsifiable test before adjudication, and tests repairs only for supported mechanisms. Existing tools automate individual pieces of this process: interpretability and auditing agents synthesize probes \citep{schwettmann2023find,shaham2024maia,li2025investigator,lu2025acd}, slice-discovery systems surface plausible weaknesses \citep{manglik2026insights,zeng2025evaltree,ashurytahan2026errormap,huang2026probellm,chung2019slicefinder,eyuboglu2022domino,kim2024b2t,chen2025hibug2,ghosh2025ladder}, and propose-and-verify systems provide statistical discipline \citep{zhong2022d3,zhong2023d5,zhou2024hypogenic,huang2025popper}. \evalrx{} joins these pieces under one decision rule: no diagnosis without statistical support, and no promoted repair without paired benefit and regression control. Each promoted repair is recorded as a \emph{verified repair recipe}: the supported diagnosis, the intervention and its tier, and the statistical evidence that justified promotion, packaged so that the fix can be audited and reapplied.

Iterative improvement is therefore a protocol, not a synonym for repeated optimization. \evalrx{} distinguishes two evolution paths. \emph{First, non-parametric evolution} (L1-L3) changes the input, surrounding scaffold, or inference-time computation while keeping checkpoint weights frozen. \emph{Second, parametric evolution} (L4) updates trainable parameters through fine-tuning or adapters and creates a new checkpoint. A repair on either path becomes version \(v_{t+1}\) only after passing the gate. That accepted version is then health-checked again for residual and newly introduced failures with fresh evidence, while the immutable root version \(v_0\) remains available to expose cumulative regression. L4 additionally requires explicit human approval.

\evalrx{} operationalizes model diagnosis and repair as a closed-loop agentic process. Starting from observed model failures, it probes the model to identify systematic failure modes (M1), analyzes candidate explanations with multiplicity-aware statistical tests (M2), formulates explicit and falsifiable hypotheses about the underlying mechanism (M3), and tests these hypotheses on the exploration split (or an optional held-out split) (M4). Supported diagnoses then guide targeted interventions (M5), whose improvements and side effects are evaluated against the unmodified model. By separating exploratory failure discovery, confirmatory diagnosis, and intervention, \evalrx{} prevents suggestive patterns from being treated as established mechanisms and proposed fixes from being treated as genuine improvements without independent validation.


\evalrx{} makes three contributions: (1) a \emph{modality-agnostic framework} that closes the loop from model probing and failure discovery to hypothesis-driven diagnosis, mechanism verification, and targeted repair. (2) a verification-gated improvement procedure spanning \emph{non-parametric} interventions at L1-L3 and \emph{parametric} adaptation at L4, where every candidate update is evaluated against both its parent and the original model, and (3) a machine-readable audit trail that captures the data splits, analysis traces, hypotheses, verification results, interventions, version transitions, and stopping decisions needed to reproduce and audit each diagnosis and repair.


%% file: sections/relatedwork.tex
\section{Related Work}
\input{sections/relatedwork_survey}

%% file: sections/relatedwork_survey.tex

\paragraph{Evaluation infrastructure and benchmark ecosystems.}
Benchmark suites and evaluation frameworks make model comparison more reproducible by standardizing datasets, prompts, metrics, and reporting. Broad, dynamic, and domain-specific evaluations complement one another: together they measure capabilities, changing task distributions, and user-facing preferences. This infrastructure is the measurement layer on which model development depends, but aggregate scores usually do not identify a coherent operational failure mechanism or specify how to repair it. \evalrx{} starts from the same labeled-evaluation setting, while treating per-case outcomes, traces, and metadata as evidence from which an actionable model-health hypothesis can be developed.

\paragraph{Discovering model failures.}
Prior work discovers model weaknesses either by actively generating probes or by mining existing evaluation results. Interpretability and auditing agents probe models to characterize mechanisms or elicit hidden behaviors~\citep{schwettmann2023find,shaham2024maia,li2025investigator,lu2025acd, kim2025agentic,huang2026probellm}, while systems such as  EvalTree and ErrorMap extract patterns or taxonomies from benchmark outputs and transcripts~\citep{manglik2026insights,zeng2025evaltree, ashurytahan2026errormap}. Related slice-discovery methods identify coherent error subpopulations using metadata, embeddings, or language-grounded attributes, primarily for vision classifiers~\citep{chung2019slicefinder, sohoni2020no,eyuboglu2022domino,kim2024b2t,chen2025hibug2, ghosh2025ladder}. \evalrx{} instead starts from a fixed labeled evaluation set for language, vision-language, or audio-language models and uses tool-executed analyses of behavioral and internal signals to form open-ended failure hypotheses.

\paragraph{Validating failure hypotheses.}
Many diagnosis systems evaluate discovered patterns using LLM judgments or heuristic consistency criteria, which do not by themselves provide false-discovery control and can yield unsupported statistical conclusions~\citep{liu2024qrdata,zhu2024statisticians,baumann2025llmhacking, asher2026phack}. Propose-and-verify methods instead test natural-language hypotheses through held-out prediction, synthetic ground truth, or sequential falsification~\citep{zhong2022d3,zhong2023d5,dunlap2025vibecheck, zhou2024hypogenic,liu2025hypobench,huang2025popper}, but have not been designed specifically for model-failure diagnosis. \evalrx{} operationalizes each failure hypothesis as an executable test, adjudicates hypotheses under a common verifier, applies BH or e-BH multiplicity correction, and permits \emph{inconclusive} outcomes~\citep{grunwald2024safetesting, wang2022ebh}.

\paragraph{Repairing model failures.}
Existing repair methods include natural-language patches, targeted data generation, model editing, and direct prompt optimization from performance feedback~\citep{murty2022patches,he2023tdg,meng2022rome, pryzant2023protegi,yuksekgonul2024textgrad,agrawal2026gepa}. Some language-grounded vision debuggers connect discovered slices to classifier retraining~\citep{kim2024b2t,chen2025hibug2,ghosh2025ladder}, whereas many LLM-diagnosis systems stop at discovery~\citep{ ashurytahan2026errormap,huang2026probellm}. In \evalrx{}, repair is instead gated by statistical validation: supported hypotheses trigger progressively deeper interventions, and only after being compared with an unchanged baseline will the most concise and effective candidate be retained.

\paragraph{Version governance and bounded recursive improvement.}
Model documentation and risk-management practices make provenance, intended use, and measured limitations inspectable. Self-improving agents add a further requirement: every promoted change can create a new distribution of failures, so a local gain is not sufficient evidence of progress. We therefore frame \evalrx{} as \emph{bounded recursive improvement}: a repair is promoted only after a baseline comparison and regression check, then the promoted version is re-audited under the same evidence discipline. The resulting audit record links the discovered issue, executable test, hypothesis verdict, intervention tier, version comparison, and promotion or rejection decision. This distinguishes a verification-gated improvement loop from open-ended optimization against self-generated feedback.

\begin{table}[H]
\centering
\small
\caption{\textbf{Where existing approaches stop, and where \evalrx{} continues.} \evalrx{} connects failure discovery, statistical hypothesis testing, repair selection, and version re-audit in one evidence-gated loop. Examples are representative, not exhaustive.}
\label{tab:related-scope}
\vspace{2pt}
\renewcommand{\arraystretch}{1.18}
\setlength{\tabcolsep}{4pt}
\begin{tabular}{>{\raggedright\arraybackslash}m{0.26\textwidth}>{\raggedright\arraybackslash}m{0.18\textwidth}>{\raggedright\arraybackslash}m{0.25\textwidth}>{\raggedright\arraybackslash}m{0.19\textwidth}}
\toprule
\tableheader
\textbf{System family \& examples} & \textbf{Starts with} & \textbf{What it delivers} & \textbf{What remains external} \\
\midrule
\tablebanding
\textbf{Auditing \& slice discovery}\newline {\scriptsize\color{evslate} EvalTree, ErrorMap} & Evaluation failures & Failure patterns or error slices & Held-out validation, repair, and post-repair audit \\
\textbf{Propose-and-verify}\newline {\scriptsize\color{evslate}D3, VibeCheck, HypoGeniC} & Candidate hypotheses & Tested hypotheses & Repair and version-level audit \\
\textbf{Prompting, editing \& retraining}\newline {\scriptsize\color{evslate}ProTeGi, TextGrad, GEPA; Claude Code$^{\ast}$} & A supplied target, reward, or task & Candidate repair & Failure diagnosis and a common verification rule \\
\rowcolor{evblue!12}\textbf{\color{evblue}\evalrx{}} & \textbf{Labeled evaluation set} & \textbf{Verified repair, or an explicit inconclusive result} & \textbf{Nothing: re-audits the promoted version and recurses} \\
\bottomrule
\end{tabular}
\par\vspace{2pt}
{\scriptsize\color{evslate}$^{\ast}$Developer agents such as Claude Code can be target-supplied repair baselines, but do not by themselves provide a verified health loop.}
\end{table}

%% file: sections/method.tex
\section{\evalrx{}: Statistically Grounded AI Diagnosis and Repair}
\label{sec:diagnostic-treatment}

\begin{figure}[t]
\centering
\includegraphics[width=\columnwidth]{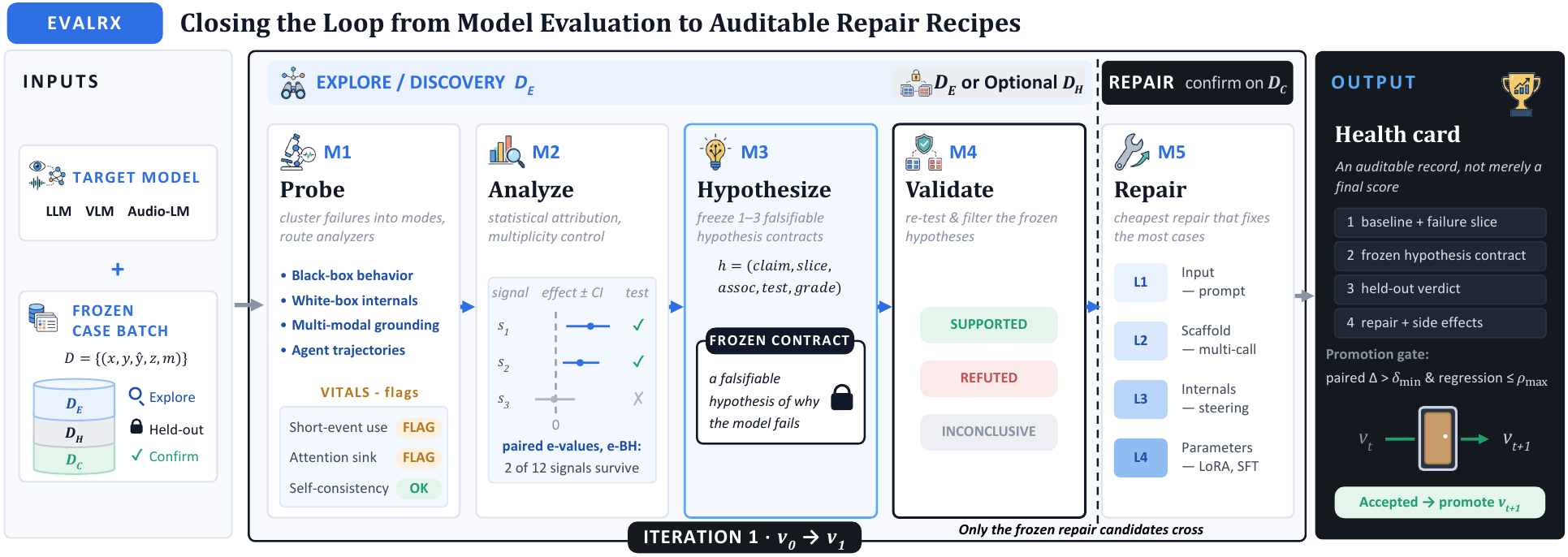}
\caption{\textbf{Overview of \evalrx{}.} \evalrx{} organizes model diagnosis and repair into five stages: \textbf{M1 (Probe)} discovers failures and gathers diagnostic evidence. \textbf{M2 (Explore)} analyzes failure structure and identifies candidate explanatory signals. \textbf{M3 (Diagnose)} formulates and freezes falsifiable hypotheses. \textbf{M4 (Verify)} tests these hypotheses with statistical evidence, and \textbf{M5 (Repair)} uses supported mechanisms to guide targeted repairs and verifies their improvement against the unchanged model.}
\label{fig:main}
\end{figure}

Figure~\ref{fig:main} gives an overview of \evalrx{}. Starting from a target model and a frozen case batch, five stages (M1–M5) move from failure discovery to verified repair. Table~\ref{tab:related-scope} positions this workflow against existing auditing, propose-and-verify, and repair systems.


\subsection{Settings: Inputs, Outputs, and Data Preprocessing}

Given a model \(f_{\theta}\) and a labeled evaluation set

$$
\mathcal{D}_0=\{(x_i,y_i,m_i)\}_{i=1}^{N},
$$

where \(x_i\) may contain text, image, audio, or multimodal inputs, \(y_i\) is the reference answer, and \(m_i\) contains permitted case metadata, \evalrx{} first executes the model under a fixed inference configuration to establish its baseline behavior. For each case, it records the model response \(\hat{y}_i=f_{\theta}(x_i)\) and a task-specific correctness indicator \(z_i\in\{0,1\}\), producing the frozen baseline case table

$$
\mathcal{D}=\{(x_i,y_i,\hat{y}_i,z_i,m_i)\}_{i=1}^{N}.
$$

Before any exploratory analysis, cases are deterministically assigned by identifier to an exploration split \(\mathcal{D}_E\),an optional sealed verification split \(\mathcal{D}_H\) and a confirmation split \(\mathcal{D}_C\) when final evaluation must be separated. The split seed, case identifiers, and any stratification variables are persisted in the run manifest, ensuring that subsequent diagnosis and repair decisions can be traced to the exact data partition on which they were made.

Only the exploration split \(\mathcal{D}_E\) is available during failure discovery, exploratory analysis, hypothesis construction, M4 adjudication and M5 candidate search. Each diagnostic hypothesis must be specified and frozen before M4 executes it (and before \(\mathcal{D}_H\) is opened when enabled). When multiple repair candidates are considered after diagnosis, the test split \(\mathcal{D}_C\) can be reserved for the final paired evaluation, preventing repair selection and final assessment from using the same cases. Throughout the process, \evalrx{} maintains a machine-readable audit trail containing the model version, data partitions, identified failure slice, frozen hypothesis specification, hypothesis verdict, repair trials, paired comparisons, and final promotion or stopping decision.

\subsection{M1: Failure Discovery and Protocol-Guided Probing}

\begin{figure}[H]
\centering
\includegraphics[width=\textwidth]{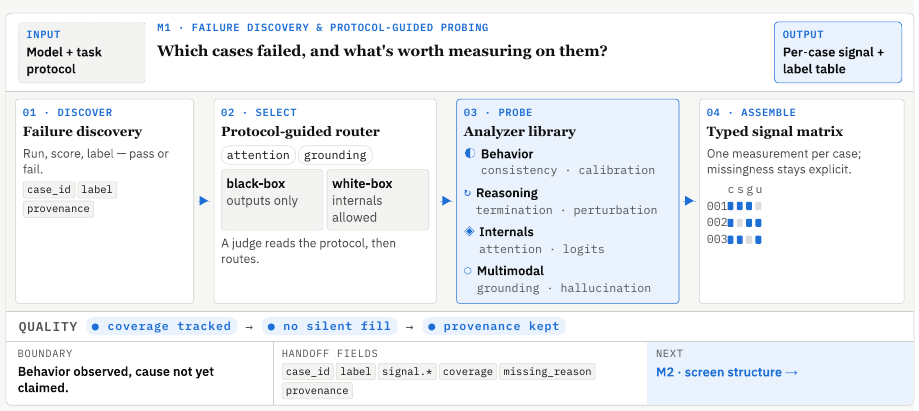}
\caption{\textbf{M1 turns raw model outcomes into an auditable diagnostic signal table.} A normalized case packet is routed to compatible analyzer families, assembled into a typed per-case signal matrix, and audited for coverage and missingness before M2.}
\label{fig:m1-flowboard}
\end{figure}

Firstly, \evalrx{} converts model behavior into structured, per-case diagnostic signals as shown in Figure~\ref{fig:m1-flowboard}. It works on individual successes and failures rather than on aggregate accuracy alone. Analyzers are selected according to the input modalities and model capabilities. The library covers black-box behavior (such as calibration, consistency, and termination), white-box signals (such as logits, attention, and representations), and multimodal failures (such as grounding errors and hallucination). Table~\ref{tab:analyzers} summarizes the analyzer families. When no built-in analyzer operationalizes a suspected mechanism, a coding agent may synthesize a diagnostic program, validate its syntax, and execute it in an isolated sandbox. These signals form the basis for subsequent statistical analysis.

\begin{table}[t]
\centering
\small
\caption{\textbf{M1 selects analyzers that match the target model and task protocol.} These representative analyzer families differ in availability across modalities and model interfaces.}
\label{tab:analyzers}
\begin{tabular}{>{\raggedright\arraybackslash}p{0.24\textwidth}>{\raggedright\arraybackslash}p{0.66\textwidth}}
\toprule
\tableheader\tablebanding
\textbf{Access and modality} & \textbf{Per-case diagnostic signals} \\
\midrule
Black-box, all modalities & Calibration, consistency under prompt or input contrasts, uncertainty, termination, response-format violations, and counterfactual outcome changes. \\
White-box language models & Token entropy, logits, attention, representation similarity, and logit-lens-style readouts. \\
Vision-language and audio-language models & Grounding, hallucination, attention or attribution over non-text tokens, modality ablations, local masking, and temporal or spatial perturbations. \\
Agentic trajectories & Loop detection, ignored-observation analysis, first-error localization, tool-use traces, and counterfactual trajectory tests. \\
\bottomrule
\end{tabular}
\end{table}

\subsection{M2-M3: Exploratory Data Analysis and Statistical Attribution with Hypothesis Generation}

\begin{figure}[H]
\centering
\includegraphics[width=\textwidth]{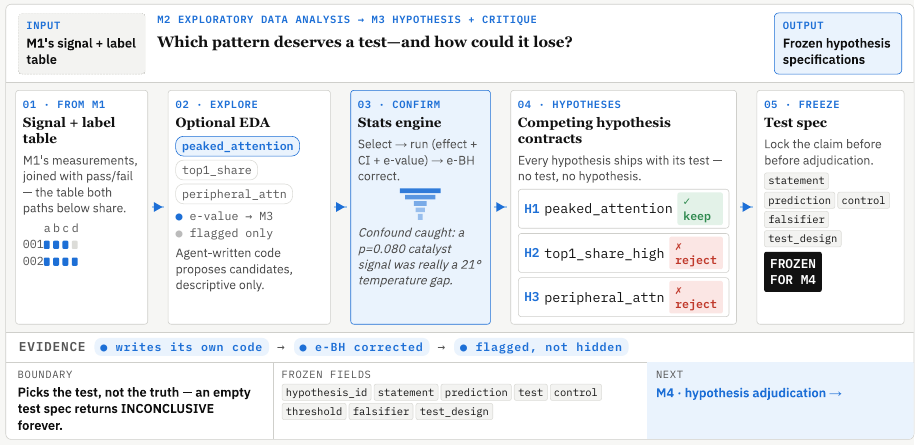}
\caption{\textbf{M2-M3 turns observed failure structure into a falsifiable, frozen test contract.} M2 records descriptive candidates; M3 challenges competing explanations and freezes the prediction, control, threshold, falsifier, and test design before M4 using them.}
\label{fig:m2-m3-flowboard}
\end{figure}

In this stage, M2 converts M1 measurements into a family of statistical analyses as shown in Figure~\ref{fig:m2-m3-flowboard}. Depending on the measurement and task, the \texttt{StatsAnalysisAgent} may use signal-label association tests, paired McNemar comparisons with e-values, bootstrap effect comparisons, rank-based tests, or single-rate tests. Every analysis records the statistic, effect estimate, confidence interval when applicable, and inferential evidence. When a run examines multiple signals or candidate modes, \evalrx{} applies BH or e-BH correction across the declared test family. In this way, \evalrx{} measures the degree to which each diagnostic signal is associated with model failure. Visualizations such as signal distributions, heatmaps, and forest plots are used for interpretation, whereas statistical tests determine which associations are considered reliable.

M3 turns selected exploratory observations from M2 into one to three falsifiable hypothesis contracts. A contract is represented as:
\[
h_k=(c_k,p_k,a_k,t_k,q_k),
\]
where \(c_k\) is a concrete mechanism claim, \(p_k\) is an executable target-slice predicate, \(a_k\) specifies the expected association and direction, \(t_k\) specifies the analyzer, metric, statistical test, and minimum effect, and \(q_k\) specifies the required evidence grade. 

This representation makes LLM-generated explanations falsifiable. For example, an assertion that a model underuses short acoustic events is not a diagnosis until it specifies which cases require a short event, which controlled perturbation should change the response, the expected direction, and the test. M3 therefore separates descriptive failure labels from claims about a mechanism.

\subsection{M4: Hypothesis Validation}

\begin{figure}[H]
\centering
\includegraphics[width=\textwidth]{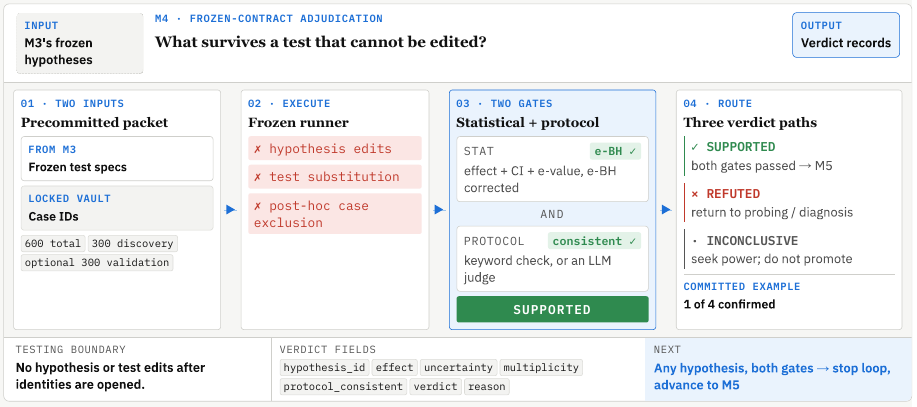}
\caption{\textbf{M4 admits a mechanism claim through statistical adjudication.} Frozen test specifications and case identifiers enter a locked runner; statistical and protocol-consistency gates jointly route each hypothesis to supported, refuted, or inconclusive.}
\label{fig:m4-flowboard}
\end{figure}

After proposing testable hypotheses, M4 executes statistical test design on \(\mathcal{D}_E\) (or optional \(\mathcal{D}_H\))(Figure~\ref{fig:m4-flowboard}). It executes the frozen contract on \(\mathcal{D}_E\)or \(\mathcal{D}_H\), applying the pre-specified analyzer in M1, statistical test in M2 and expected direction in M3. A hypothesis is \emph{supported} only when the effect has the expected direction, meets the declared minimum effect, survives the specified multiplicity procedure, and satisfies the task protocol. It is \emph{refuted} when the evidence contradicts the prediction and \emph{inconclusive} otherwise.

\evalrx{} additionally distinguishes observational from intervention-grade evidence. An association between a naturally occurring signal and error is observational; a paired strategy contrast, controlled perturbation, or targeted mechanism test can provide intervention-grade evidence. 

\subsection{M5: Mechanism Intervention, Repair, and Verification}



Lastly, validated mechanisms are converted into interventions using a repair ladder ordered by implementation cost and invasiveness. The current repair ladder organization is shown in Table~\ref{tab:repair-ladder}, the cost increases sequentially from L1 to L4.

\begin{table}[t]
\centering
\small
\caption{\textbf{The repair ladder makes intervention cost and parameter change explicit.} L1-L3 are non-parametric evolution because the checkpoint weights remain frozen; L4 is parametric evolution. The tier constrains the intervention space, not repair quality.}
\label{tab:repair-ladder}
\begin{tabular}{>{\raggedright\arraybackslash}p{0.07\textwidth}>{\raggedright\arraybackslash}p{0.17\textwidth}>{\raggedright\arraybackslash}p{0.14\textwidth}>{\raggedright\arraybackslash}p{0.48\textwidth}}
\toprule
\tableheader\tablebanding
\textbf{Tier} & \textbf{Evolution path} & \textbf{Space} & \textbf{Candidate treatment} \\
\midrule
L1 & Non-parametric & Input & Prompt rewrites, explicit constraints, output formats, and evidence-use instructions. \\
L2 & Non-parametric & Scaffold & Deterministic multi-call workflows, preprocessing, external tools, conditional execution, and output aggregation around the unchanged model. \\
L3 & Non-parametric & Internals & Read-only logits or attention, or audited inference-time interventions such as attention reweighting or activation steering, without persistent weight updates. \\
L4 & Parametric & Parameters & Dataset construction and fine-tuning or adapter updates. These create a new checkpoint and require explicit human execution or approval. \\
\bottomrule
\end{tabular}
\end{table}

The framework begins with the lowest repair tier capable of addressing the validated mechanism and escalates when lower-cost interventions are ineffective. 

\begin{figure}[H]
\centering
\includegraphics[width=\textwidth]{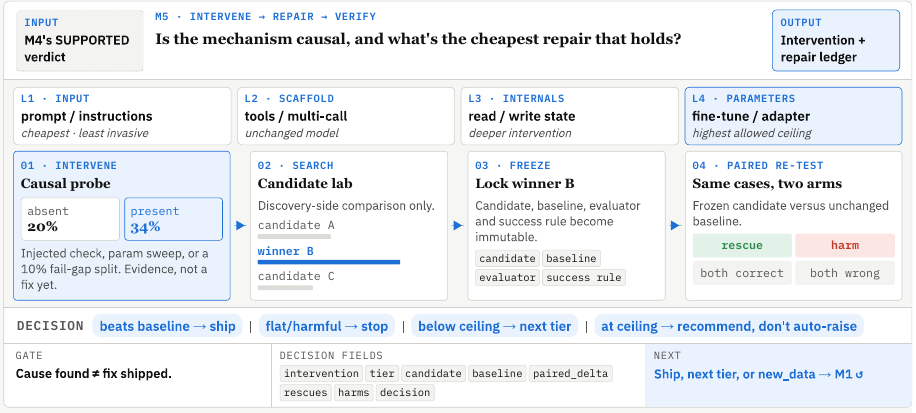}
\caption{\textbf{M5 promotes only repairs that beat the unchanged baseline under a paired test.} A supported hypothesis first receives a causal probe; repair search then starts at the least invasive permitted tier, freezes a candidate and its success rule, and compares it with the unchanged baseline on matched cases. The decision is to ship, stop, try the next tier, or return filtered data to M1.}
\label{fig:m5-flowboard}
\end{figure}

Repair candidates are search runs on the exploration split \(\mathcal{D}_E\) (or optional \(\mathcal{D}_H\)). 
Candidates from the lowest permitted tier are compared with the baseline on the same exploration cases. Then the best candidate and its success rule are frozen. The frozen candidate is then compared with the unchanged baseline through paired evaluation on identical cases drawn from the sealed confirmation split \(\mathcal{D}_C\)(Figure~\ref{fig:m5-flowboard}).
A repair is accepted only when its net effect satisfies the pre-specified improvement criterion and does not introduce unacceptable degradation. An accepted intervention is then persisted, together with its supported diagnosis, tier, and paired evidence, as a \emph{verified repair recipe} that can be reapplied to new cases or audited later.



\subsection{Verified Evolution and Iterative Improvement}

\textbf{EvalRx makes iterative improvement a sequence of auditable version decisions, not unconstrained self-modification.} At iteration \(t\), the system begins with the currently promoted version \(v_t\), a frozen diagnostic hypothesis \(h_t\), and a candidate repair \(r_t\). It promotes the candidate only when the diagnosis is supported by its frozen test, the repair improves the affected behavior in a held-out paired comparison, regressions remain within a predeclared bound, and the intervention stays within the approved repair ceiling:
\[
A_t=
\mathbf{1}[\operatorname{Verify}(h_t)]
\mathbf{1}[\Delta_{\mathrm{target}}(r_t;v_t)>\delta_{\min}]
\mathbf{1}[R_{\mathrm{non\mbox{-}target}}(r_t;v_t,v_0)\leq\rho_{\max}]
\mathbf{1}[\operatorname{tier}(r_t)\leq L_{\max}].
\]
Here, \(v_0\) is the immutable root version and serving configuration, \(\delta_{\min}\) is the minimum paired target benefit, \(\rho_{\max}\) is the tolerated non-target regression, and \(L_{\max}\) is the approved intervention ceiling. Statistical evidence requirements, including significance or e-value criteria where applicable, are part of the verification and paired-benefit predicates.

The gate defines a conservative version transition:
\[
v_{t+1}=
\begin{cases}
r_t(v_t), & A_t=1,\\
v_t, & A_t=0.
\end{cases}
\]
An unsuccessful candidate therefore remains an auditable result rather than silently changing the deployed model. The immutable root \(v_0\) remains available for every later comparison, so a chain of local gains cannot conceal accumulated regression.

\textbf{The repair tier determines which kind of model evolution is permitted.} L1-L3 are \emph{non-parametric}: they change prompts, scaffolds, tools, or inference-time computation while checkpoint weights remain frozen. L4 is \emph{parametric}: it creates a separately versioned adapter or checkpoint through fine-tuning. Both paths must pass the same hypothesis, paired-benefit, and regression gates; L4 additionally requires explicit human approval before a new parameterized artifact can be promoted.

\textbf{Recursion occurs only after a supported promotion and fresh evidence allocation.} Once \(v_{t+1}\) is promoted, it may enter a new M1-M5 cycle on a newly declared evaluation split or batch. A failed candidate instead records a rejected edge, stops, or escalates within \(L_{\max}\). This is the operational sense in which \evalrx{} supports iterative model improvement: each iteration preserves the evidence, version lineage, and stopping reason required to audit the next one.
\begin{algorithm}[H]
\caption{\textbf{A guarded iterative version loop.} A version can seed the next cycle only after frozen-contract diagnosis, held-out paired repair evaluation, and the approved intervention ceiling all pass.}
\label{alg:iterative}
\begin{tcolorbox}[colback=evblue!3,colframe=evblue!65!black,boxrule=0.8pt,arc=2pt,left=8pt,right=8pt,top=6pt,bottom=6pt]
\small
\textbf{\color{evblue}INPUT}\quad immutable root \(v_0\), fresh evaluation stream \(\{\mathcal{D}^{(t)}\}\), candidate budget \(B\), tier ceiling \(L_{\max}\), and stop policy.\par
\vspace{2pt}
\textbf{\color{evblue}INITIALIZE}\quad \(t\leftarrow0\), promoted version \(v_t\leftarrow v_0\).
\begin{enumerate}[leftmargin=1.7em,itemsep=2pt,topsep=5pt]
\item \textbf{Diagnose (M1-M2).} Probe \(v_t\) on exploration cases and record failure modes plus diagnostic signals.
\item \textbf{Precommit (M3).} Freeze executable hypothesis contracts before adjudication.
\item \textbf{Verify (M4).} Test each contract against its declared effect, direction, and multiplicity family.
\item \textbf{Repair and compare (M5).} For a supported contract, search at most \(B\) candidates from the lowest permitted tier; compare target benefit and regressions with \(v_t\) and \(v_0\).
\item \textbf{Decide.} If \(A_t=1\), record \(v_t\xrightarrow{r_t}v_{t+1}\), allocate fresh evidence, and return to Step~1. Otherwise record the rejected edge and stop or escalate only within \(L_{\max}\).
\end{enumerate}
\vspace{1pt}
\textbf{\color{evblue}OUTPUT}\quad version graph, health-card registry, rejected candidates, and explicit stopping reason.
\end{tcolorbox}
\end{algorithm}

Algorithm~\ref{alg:iterative} makes the iterative claim testable. The original baseline $v_0$ remains available for every later paired comparison, preventing a sequence of local gains from obscuring cumulative regression. A repair must be tested on cases not used to select it; when an iteration consumes its held-out split, the next iteration uses a fresh, pre-declared split or new evaluation batch. The intervention ceiling is carried forward, so the system may search within the non-parametric L1-L3 path but cannot silently cross into parametric L4 evolution. L4 parameter updates, ceiling increases, and deployment decisions require explicit human approval and a separately versioned checkpoint artifact.

The loop halts when no supported failure mode remains under the declared scope, when the next candidate is ineffective or regressive, when the intervention ceiling or resource budget is reached, or when a human stops the investigation. Thus, iterative improvement produces a version graph of supported and rejected changes, not an unconstrained self-modifying system. The health-card registry reports each node, edge, evidence split, and stop reason.



%% file: sections/setup.tex
\section{Evaluation Scope and Setup}
\subsection{Task and Model Roster}

The evaluation is designed to test whether the same evidence-to-repair protocol transfers across three distinct input modalities: Large Language Models (LLMs), Vision-Language Models (VLMs), and Audio-Language Models (ALMs). Each track uses two to four predeclared datasets and a fixed model roster across both datasets.
For visual tasks, we select ChartQA~\cite{masry2022chartqa}, Spatial457~\cite{wang2025spatial457} and Chair~\cite{rohrbach2018object} to diagnose model's grounding and hallucination problems. As language reasoning tasks, we choose two BBH subtasks (word sorting and causal judgement)~\cite{suzgun2023challenging}, GSM8K~\cite{cobbe2021trainingverifierssolvemath}, and HotpotQA~\cite{yang2018hotpotqa}, covering algorithm execution, causal judgement, mathematical reasoning, and multi-hop question answering. ALMs track targets environmental-sound understanding, general audio reasoning, and music understanding using MMAU~\cite{sakshi2025mmau} and AudioCaps~\cite{kim2019audiocaps} datasets. Dataset versions, scoring rules, split seeds, and serving parameters are frozen in the run manifest before diagnosis begins. As shown in Table~\ref{tab:modelset}, we use both closed-source model Gemini and open-weight models including Qwen, Gemma and Nemotron ranging from 2B to 30B parameters as evaluation targets.

\begin{table}[H]
\centering
\small
\caption{\textbf{A three-modality evaluation roster.} The table lists the models in benchmark and the role of each LLM, VLM, and ALM track in evaluating \evalrx{}.}
\label{tab:modelset}
\begin{tabular}{>{\raggedright\arraybackslash}p{0.14\textwidth}>{\raggedright\arraybackslash}p{0.50\textwidth}>{\raggedright\arraybackslash}p{0.26\textwidth}}
\toprule
\tableheader\tablebanding
Task & Models & Role \\
\midrule
LLM & Gemini-2.5-Flash-Lite, Qwen3.5 2B, Gemma E2B,\newline Nemotron-3-Nano-4B & Four text datasets, cross-scale and cross-family comparison. \\
VLM & Gemini-2.5-Flash-Lite, Qwen3.5 2B, Gemma E2B;\newline Nemotron-3-Nano-Omni-30B-A3B-Reasoning-FP8 & Three visual datasets, cross-scale and cross-family comparison. \\
ALM & Gemini-2.5-Flash-Lite, Qwen3-Omni-30B-A3B-Instruct,\newline Gemma E2B,\newline Nemotron-3-Nano-Omni-30B-A3B-Reasoning-FP8 & Two audio datasets, cross-family comparison. \\
\bottomrule
\end{tabular}
\end{table}

\subsection{Serving and Inference Configuration}
The evaluation targets are the models in Table~\ref{tab:modelset}; diagnostic or coding agents are auxiliary pipeline components, not additional models in the comparison. \evalrx{} records the identity and version of every auxiliary agent, scorer, analyzer, prompt, decoding setting, and serving configuration in the run manifest. This separation makes the unit of a health-card conclusion unambiguous: a specific model revision under a specific serving configuration on a specific dataset.

For the initial pipeline configuration, we use Claude Code with Opus-5 as the base orchestration agent and additional Antigravity with Gemini-3.1-Pro to ensure that the system works with different kinds of agents. 





%% file: sections/experiment.tex
\section{Empirical Evaluation}
\label{sec:evaluation}

\subsection{Quantitative Results}
We evaluate \evalrx{} as a diagnosis-and-repair system rather than as a score-optimization procedure. The quantitative study asks whether verified repairs improve the affected behavior across model families, modalities, and agent backbones.

\begin{figure}[H]
    \centering
    \includegraphics[width=0.9\linewidth]{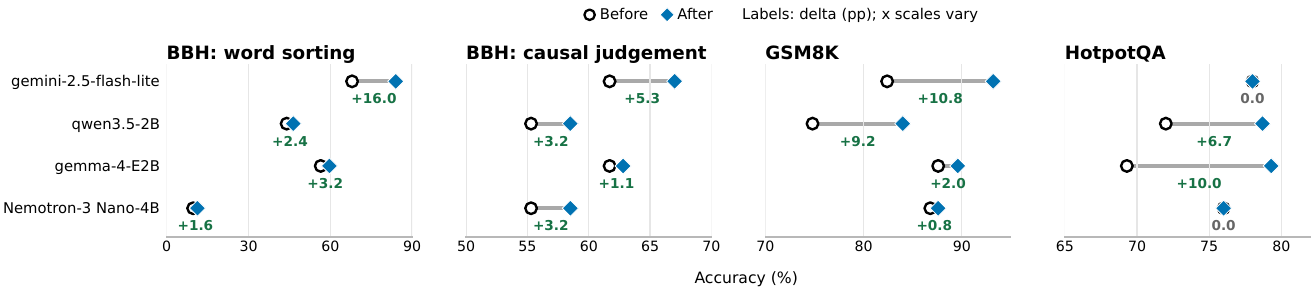}
    \caption{\textbf{Verified repairs across the LLM benchmark track.} Each panel compares baseline and post-repair performance; labels report the panel-specific change in percentage points.}
    \label{fig:llm_bench_claude}
\end{figure}

\paragraph{LLM experiments across families.} Figure~\ref{fig:llm_bench_claude} summarizes the improvement for the LLM track across four benchmarks. Overall, accuracy improved in 14 of the 16 model–task combinations. The average improvement is 4.7 percentage points (pp), although the magnitude of the gains varied substantially across tasks. The largest individual improvements are observed for Gemini-2.5-Flash-Lite on word sorting (+16.0 pp) and GSM8K (+10.8 pp), and Gemma-4-E2B on HotpotQA (+10.0 pp). Moreover, Qwen3.5-2B and Gemma-4-E2B improve consistently across all four benchmarks, demonstrating the general applicability of \evalrx{}.

\begin{figure}[H]
    \centering
    \includegraphics[width=0.8\linewidth]{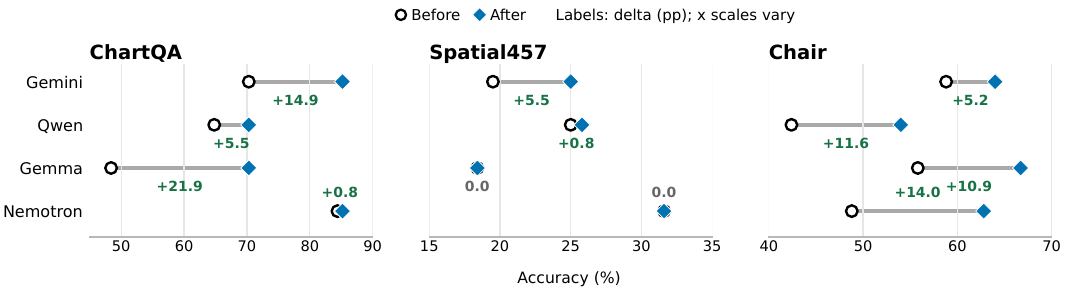}
    \caption{\textbf{Verified repairs across the VLM benchmark track.} The figure compares baseline and post-repair performance using Claude Opus-5 as the base agent.}
    \label{fig:vlm_bench_claude}
\end{figure}

\paragraph{VLM experiments across families.}
Figure~\ref{fig:vlm_bench_claude} reports the accuracy of four vision-language models before and after applying \evalrx{}. Performance improves in 10 of the 12 model–task combinations. The mean improvement across all settings is 7.6\%. However, the gains are strongly task-dependent: the average improvements on ChartQA and Chair were 10.8 pp and 10.4 pp, respectively, whereas Spatial457 shows an average gain of only 1.6 pp. Gemini and Qwen improve on all three benchmarks, while Gemma and Nemotron remain unchanged on Spatial457.

\begin{figure}[H]
    \centering
    \includegraphics[width=0.6\linewidth]{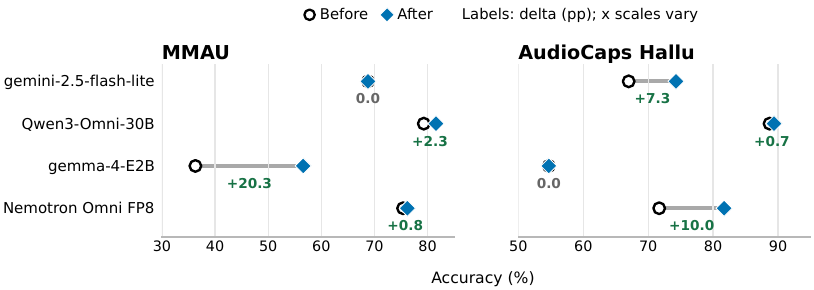}
    \caption{\textbf{Verified repairs across the ALM benchmark track.} The figure compares baseline and post-repair performance using Claude Opus-5 as the base agent.}
    \label{fig:alm_bench_claude}
\end{figure}

\paragraph{ALM experiments across families.}
Figure~\ref{fig:alm_bench_claude} reports the corresponding audio-language results under the same agent configuration. Performance improves in six of the eight model–benchmark combinations. These results suggest that the effect of \evalrx{} is strongly dependent on both the underlying model and the evaluated capability. In particular, Gemma exhibits a substantial improvement in general audio understanding on MMAU but no improvement in audio hallucination detection. Nemotron shows the opposite pattern, obtaining only a small gain on MMAU but a considerably larger improvement on AudioCaps. This complementary behavior indicates that \evalrx{} addresses different model-specific failure modes rather than providing a uniform improvement in audio-language understanding. 

\begin{figure}[H]
    \centering
    \includegraphics[width=0.9\linewidth]{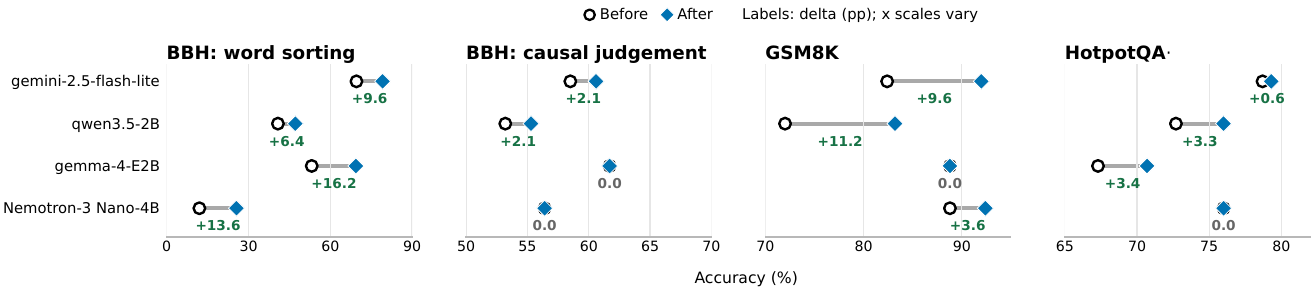}
    \caption{\textbf{LLM repair outcomes with a second agent backbone.} The figure compares baseline and post-repair performance using Antigravity Gemini-3.1-Pro as the base agent.}
    \label{fig:llm_bench_agy}
\end{figure}

\paragraph{Additional agent-backbone experiment.}
Figure~\ref{fig:llm_bench_agy} repeats the LLM comparison with Antigravity (Gemini-3.1-Pro) as the base agent. Accuracy improves in 12 of the 16 model–task combinations, with an average gain of 5.1 pp; the largest improvements are observed for Gemma-4-E2B (+16.2 pp) and Nemotron-3-Nano-4B (+13.6 pp) on word sorting, and Qwen3.5-2B on GSM8K (+11.2 pp). Similar to using Claude as the base agent, Antigravity can also effectively repair potential issues in the model in most cases.

\paragraph{Effectiveness across model scales within the Gemma-4 family.}
\begin{table*}[t]
\centering
\caption{\textbf{Repair gains across the Gemma-4 scale range.} The table reports baseline accuracy and post-repair improvement for each Gemma-4 model-task setting.}
\label{tab:frozen-repair-results}
\resizebox{\textwidth}{!}{%
\begin{tabular}{lcccccc}
\toprule
\textbf{Model}
& \textbf{BBH (word sorting)}
& \textbf{gsm8k}
& \textbf{MMAU}
& \textbf{AudioCaps-Hallucination}
& \textbf{ChartQA}
& \textbf{Spatial457} \\
\midrule

Gemma-4-E2B
& $56.5\%$ \improve{3.2}
& $87.6\%$ \improve{2.0}
& $36.3\%$ \improve{20.3}
& $56.6\%$ \improve{0.0}
& $48.4\%$ \improve{21.9}
& $18.4\%$ \improve{0.0} \\

Gemma-4-E4B
& $62.0\%$ \improve{21.6}
& $91.4\%$ \improve{0.0}
& $65.3\%$ \improve{12.0}
& $61.7\%$ \improve{16.7}
& $58.3\%$ \improve{26.0}
& $16.7\%$ \improve{22.7} \\

Gemma-4-12B
& $82.4\%$ \improve{11.2}
& $93.4\%$ \improve{1.2}
& $63.0\%$ \improve{16.7}
& $54.0\%$ \improve{18.7}
& $73.7\%$ \improve{17.3}
& $16.3\%$ \improve{19.3} \\

\bottomrule
\end{tabular}%
}
\end{table*}

We evaluate two datasets each for vision-language, text-only language and audio-language tasks using Gemma-4 models at 2B, 4B, and 12B parameters. This setting tests whether the workflow transfers across model scale rather than being tied to one checkpoint.

As shown in Table~\ref{tab:frozen-repair-results}, \evalrx{} is effective across all three model sizes, yielding average accuracy gains of 7.9, 16.5, and 14.1 percentage points for the E2B, E4B, and 12B models, respectively. Performance improves for most model–dataset pairs, with three settings remaining unchanged: E4B on GSM8K and E2B on AudioCaps-Hallucination and Spatial457. The gains are particularly substantial on the multimodal benchmarks, reaching up to 26.0 percentage points. The smaller improvements on GSM8K are likely due to its already high baseline accuracy, which leaves limited room for further improvement. Overall, these results suggest that the effectiveness of \evalrx{} is not tied to a particular model scale and generalizes across models with different capacities.

\subsection{Case Studies}
The following traces make the diagnosis-to-repair process inspectable. They report the discovered behavior, hypothesis verdict, the selected repair tier, and the paired case-level outcome. Crucially, the three cases have different evidentiary status: ChartQA and BBH word sorting demonstrate mechanism-verified repairs, whereas MMAU demonstrates a measured repair whose mechanism remains unresolved. We retain both outcomes because the system should not turn an accuracy gain into an unsupported causal explanation. Because the content generated using Claude as the base agent has poor readability, we use Antigravity as the base agent in the qualitative study. 


\begin{figure}[H]
    \centering
    \includegraphics[width=\linewidth]{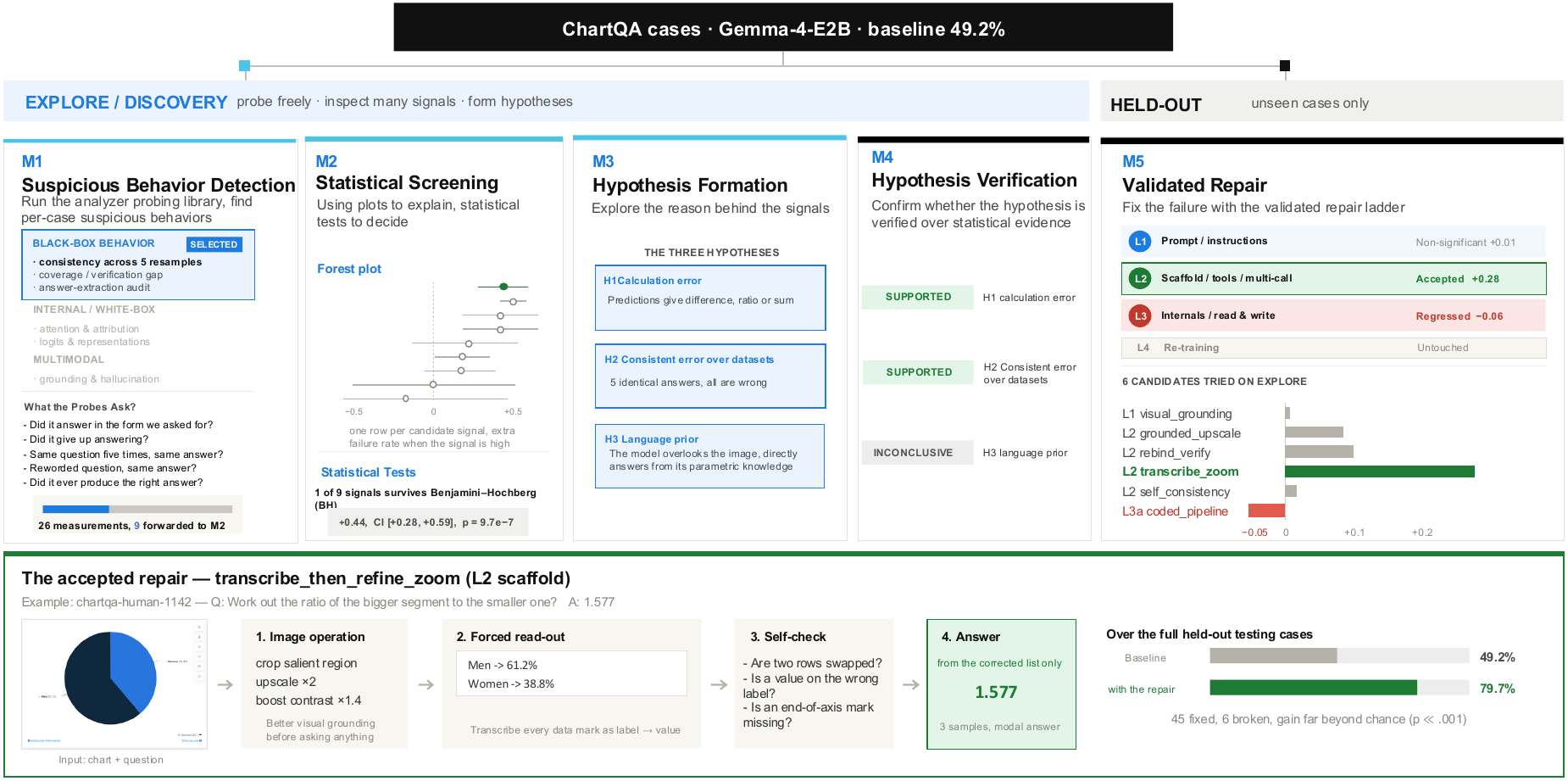}
    \caption{\textbf{A complete, mechanism-verified repair loop on ChartQA.} For Gemma-4-E2B, the L2 scaffold is promoted only after its hypotheses are supported under frozen tests and its paired outcome improves from $49.2\%$ to $79.7\%$.}
    \label{fig:qual_vlm}
\end{figure}

\paragraph{ChartQA: a complete evidence-to-repair loop.}
Figure~\ref{fig:qual_vlm} is the positive end-to-end example. From 26 measurements, the discovery stage forwards nine signals for screening; one survives the Benjamini-Hochberg gate ($\Delta=0.44$, CI $[0.28,0.59]$, $p=9.7\mathrm{e}{-7}$). The resulting calculation-error and cross-dataset-consistency hypotheses are supported under frozen tests. The repair search then keeps the least invasive successful intervention: the L2 \emph{transcribe then refine zoom} scaffold. On the held-out evaluation, accuracy rises from $49.2\%$ to $79.7\%$, with 45 cases fixed and six broken. This trace is the intended verification-gated loop: the intervention is promoted together with the evidence that supports its diagnosis and bounds its regressions.

\begin{figure}[H]
    \centering
    \includegraphics[width=\linewidth]{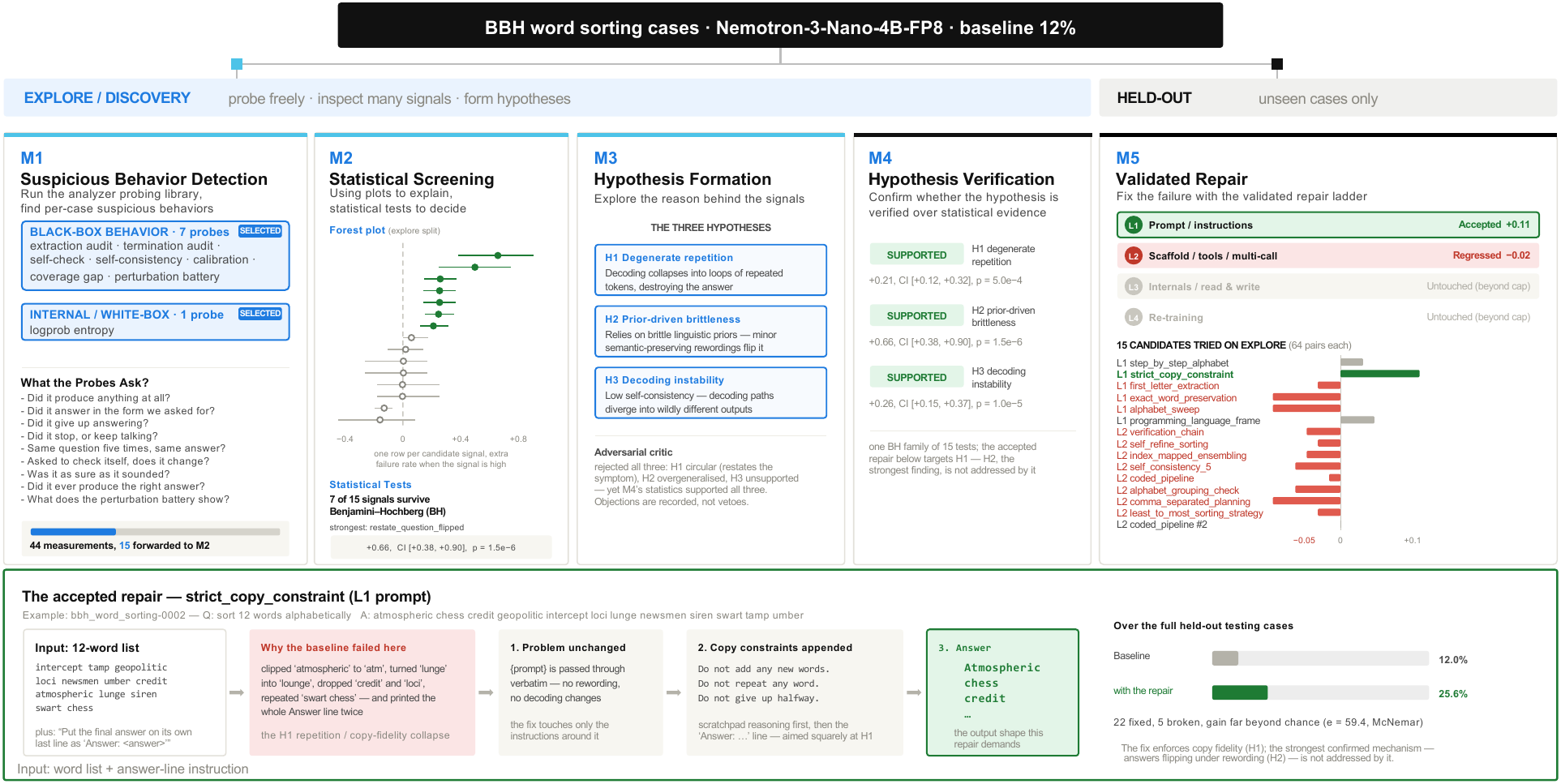}
    \caption{\textbf{A mechanism-verified L1 repair on BBH (word sorting).} For Nemotron-3-Nano-4B-FP8, the supported strict-copy intervention improves held-out accuracy from $12.0\%$ to $25.6\%$.}
    \label{fig:qual_llm}
\end{figure}

\paragraph{BBH (word sorting): a verified repair with prompt optimization.} 
Figure~\ref{fig:qual_llm} presents a verified repair. From 44 measurements, the discovery stage forwards 15 signals for screening, of which seven survive the Benjamini-Hochberg gate. The strongest finding is sensitivity to semantic-preserving restatement ($\Delta=0.66$, CI $[0.38,0.90]$, $p=1.5\mathrm{e}{-6}$). Adjudication supports all three resulting hypotheses: degenerate repetition, prior-driven brittleness, and decoding instability. The repair search then retains the least invasive successful intervention, the L1 \emph{strict copy constraint} prompt, which directly targets the supported repetition and copy-fidelity failure. On held-out cases, accuracy increases from $12.0\%$ to $25.6\%$, with 22 cases fixed and five broken.

\begin{figure}[H]
    \centering
    \includegraphics[width=\linewidth]{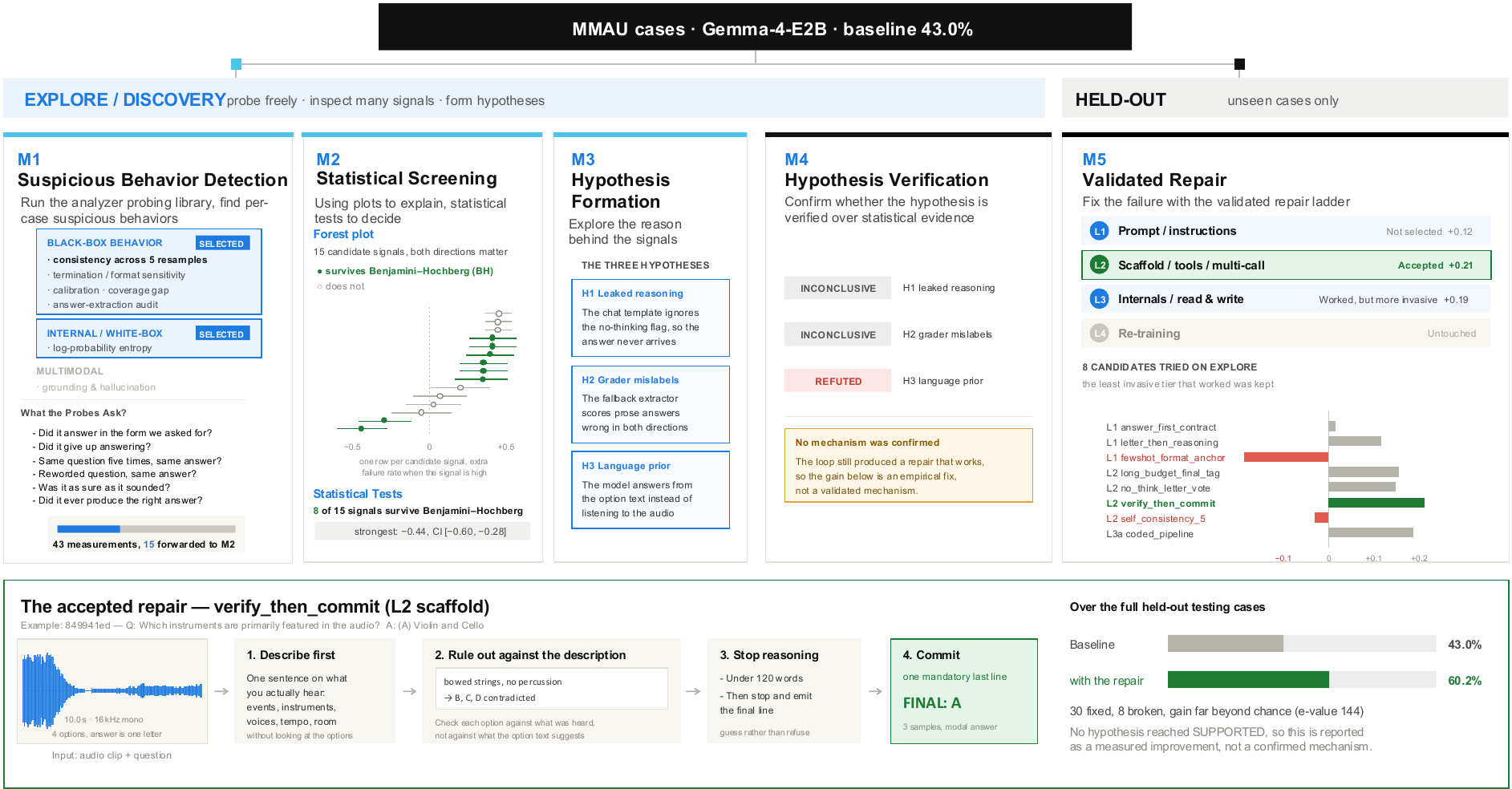}
    \caption{\textbf{An empirical repair that remains mechanistically unresolved on MMAU.} For Gemma-4-E2B, the L2 scaffold improves held-out accuracy from $43.0\%$ to $60.2\%$, but no mechanism reaches the supported verdict; the result is therefore not presented as a confirmed mechanism repair.}
    \label{fig:qual_alm}
\end{figure}

\paragraph{MMAU: an empirical repair pending diagnosis.}
Figure~\ref{fig:qual_alm} illustrates the complementary boundary case. The system detects and screens candidate signals, but no hypothesis reaches the supported verdict: two remain inconclusive, and the language-prior hypothesis is refuted. So M5 stage can only take inconclusive hypotheses into consideration. Nevertheless, inspired by inconclusive hypotheses, the L2 \emph{verify then commit} scaffold produces a measured held-out improvement from $43.0\%$ to $60.2\%$, with 30 cases fixed and eight broken. We therefore report this result as an empirical repair rather than a mechanism-verified closed-loop repair. Its value is operational and diagnostic: it identifies a useful intervention while preserving the speculative mechanism, the rejected alternatives, and the path for a subsequent investigation.

\clearpage

%% file: sections/conclusion.tex
\section{Discussion and Future Directions}

\evalrx{} treats a low benchmark score as the beginning of an investigation rather than its conclusion. Its unit of analysis is an auditable health card: a reproducible failure slice, a frozen mechanism hypothesis, hypothesis verdict, held-out paired evidence, a repair attempt, and a paired account of benefits and regressions. This makes negative outcomes useful. A candidate may fail because the discovered pattern does not generalize, the mechanism is unsupported, the repair is ineffective, or the repair harms previously correct cases; these are scientifically different conclusions and should remain distinguishable.

The current study has clear limitations. It does not claim that every failure mode is repairable or that every statistically supported association establishes a complete causal explanation. Public benchmark contamination, finite probe coverage, imperfect grading of free-form outputs, serving-stack dependence, and incomplete access to model internals limit the conclusions that can be drawn. The multiple datasets design in each modality provides a regression panel, but it is not a substitute for a comprehensive safety evaluation or deployment assessment.

The release is designed to make these limitations inspectable. It will include case manifests, split definitions, model and serving configurations, prompt and configuration hashes, analyzer traces, frozen hypothesis contracts, raw result records, repair trials, and failed reproductions. These artifacts allow another group to reproduce a reported health card, challenge its mechanism claim, or test a new repair without changing the evaluation contract. A health card is evidence for technical review and monitoring; it is not, by itself, a claim of safety certification or regulatory compliance.

\paragraph{Trade-offs of an optional validation set.}
Although a cleaner design would use separate validation split for M4 hypothesis verification and M5 repair candidates selection, we don't introduce an additional held-out split in the present experiments.
The validation sets contain relatively few examples from each discovered failure slice, making it difficult for genuine effects to reach the prespecified evidentiary thresholds.
In pilot runs, M4 verification in held-out data already limits the number of supported mechanisms available to guide repair generation because of limited samples. Using the same small validation set for candidate selection would introduce an additional low-power decision: a candidate would first need to outperform alternatives on validation data and then demonstrate improvement again on the final test data.
Candidate rankings on such small validation samples can be dominated by sampling variability, even when the candidate pool is unchanged.
We therefore select repair candidates using the exploration data and reserve unseen test cases for final confirmation. 
This choice improves selection power but may favor interventions tailored to the diagnosed examples. With larger validation sets, held-out repair selection would be preferable because it would provide stronger protection against adaptive overfitting.

\paragraph{Toward recursive self-improvement.}
The next frontier is to turn individual repair recipes into \emph{cumulative, evidence-guided model improvement}. The present study evaluates at most one promoted repair per model--task setting; sustained improvement across generations remains an empirical question. Building on Algorithm~\ref{alg:iterative}, future work will repeat the M1--M5 cycle on each promoted version, using fresh cases for final repair confirmation and paired comparisons against both the parent version and the immutable root \(v_0\). The key question is whether successive repairs compose: do gains accumulate, plateau as the remaining failures become harder, or interfere with behaviors repaired earlier? These experiments will also examine when non-parametric interventions at L1--L3 should give way to human-approved L4 adaptation, tracking cumulative accuracy, regressions, and inference cost rather than isolated score gains.

A second direction is to make repair recipes reusable across tasks and model versions. A recipe registry could support retrieval and adaptation of previously successful interventions, while requiring fresh evaluation before a transferred repair is accepted. Comparing diagnosis-guided search with direct repair search under matched agent and compute budgets would establish when explicit diagnosis improves repair quality or search efficiency. This would turn the audit trail into a resource for learning which failure signatures predict useful interventions, without assuming that a successful repair establishes its proposed mechanism.

The longer-term ambition is \emph{recursive self-improvement of the diagnostic system itself}. Beyond repairing a target model, \evalrx{} could use recorded successes and failures to propose changes to its analyzers, hypothesis-generation policies, and repair-search strategies. Such changes should be evaluated on held-out model--task settings under an externally fixed acceptance protocol, rather than allowing the improving system to redefine its own success criterion. The goal is a system that becomes better at finding and repairing failures over time, with each accepted change remaining traceable to an executable intervention and independently evaluated benefit.


\section{Conclusion}

We introduce \evalrx{}, an agentic system that turns model evaluation into statistically grounded model diagnosis and repair. By separating exploratory discovery from frozen hypothesis adjudication and held-out paired repair verification, \evalrx{} turns model errors into measurable signals, statistically grounded hypotheses, and verified repair recipes.
By separating exploration from final test and evaluating repairs against an unchanged baseline, it provides a systematic and reproducible path from failure discovery to model improvement. 
The benchmark tests this workflow across Gemini, Qwen, Gemma, and Nemotron models, with two to four datasets in each modality track. The central outcome is not a larger score alone, but a traceable answer to which behavior failed, what evidence supports the diagnosis, whether the proposed repair generalized beyond the cases that motivated it and which fix level path was taken to repair model while saving costs.